\documentclass[letterpaper, 10 pt, conference]{ieeeconf}  

\IEEEoverridecommandlockouts                              

\usepackage{graphicx} 
\usepackage{amsmath} 
\usepackage{algorithm}
\usepackage{algpseudocode}
\usepackage{dblfloatfix}
\usepackage[absolute]{textpos}
\usepackage[some,bottom]{background}

\title{\LARGE \bf
E2R: a Hierarchical-Learning inspired Novelty-Search method to generate diverse repertoires of grasping trajectories
}

\author{Johann Huber$^{1}$, Oumar Sane$^{1}$, Alex Coninx$^{1}$, Faïz Ben Amar$^{1}$ and Stéphane Doncieux$^{1}$ 
\thanks{$^{1}$Sorbonne Université, CNRS, Institut des Systèmes Intelligents et de Robotique, ISIR, F-75005 Paris, France {\tt\small \{huber, sane, coninx, amar, doncieux\}@isir.upmc.fr}}%
}

\SetBgContents{This work has been submitted to the IEEE for possible publication. Copyright may be transferred without notice, after which this version may no longer be accessible.}
\SetBgScale{0.8}
\SetBgOpacity{1}
\SetBgColor{gray}
\SetBgVshift{1cm}

\begin{document}



\maketitle
\thispagestyle{empty}
\pagestyle{empty}

\begin{abstract}
Robotics grasping refers to the task of making a robotic system pick an object by applying forces and torques on its surface. Despite the recent advances in data-driven approaches, grasping remains an unsolved problem. Most of the works on this task are relying on priors and heavy constraints to avoid the exploration problem. Novelty Search (NS) refers to evolutionary algorithms that replace selection of best performing individuals with selection of the most novel ones. Such methods have already shown promising results on hard exploration problems. In this work, we introduce a new NS-based method that can generate large datasets of grasping trajectories in a platform-agnostic manner. Inspired by the hierarchical learning paradigm, our method decouples approach and prehension to make the behavioral space smoother. Experiments conducted on 3 different robot-gripper setups and on several standard objects shows that our method outperforms state-of-the-art for generating diverse repertoire of grasping trajectories, getting a higher successful run ratio, as well as a better diversity for both approach and prehension. Some of the generated solutions have been successfully deployed on a real robot, showing the exploitability of the obtained repertoires.  

\end{abstract}


\section{INTRODUCTION}

Robotics grasping is often considered as a prerequisite for manipulation skills \cite{hodson2018gripping}, which has lead to great interest in the past decades \cite{kleeberger2020survey}. Recent progress in the field of Machine Learning resulted into significant results on the task
\cite{levine2018learning}\cite{morrison2020egad}. However, its sparse-reward nature make Reinforcement Learning (RL) algorithms \cite{sutton2018reinforcement} ineffective without additional engineering: the learning agent cannot easily distinguish promising rollouts from unpromising ones, as any attempt results into a null reward as long as the task is not fulfilled.

Many solutions has been proposed to avoid the exploration problem: leveraging reward shaping \cite{ng1999policy}, bootstrapping the optimisation process with demonstrations \cite{argall2009survey} or relying on heuristic primitives \cite{levine2018learning}. However, all those approaches impose strong constraints on the learned policies: reward shaping can push the policy toward the maximisation of sub-objectives which might not be aligned with the task, Imitation Learning algorithms are known to struggle as soon as the agent deviates from the learning distribution \cite{hussein2017imitation}, while the heuristic policies implies constraining both the trajectories and the operational space to make rough hand-crafted controllers able to produce successful grasps before the learning starts.

\begin{figure}[btp]
  \centering
  \includegraphics[width=\linewidth]{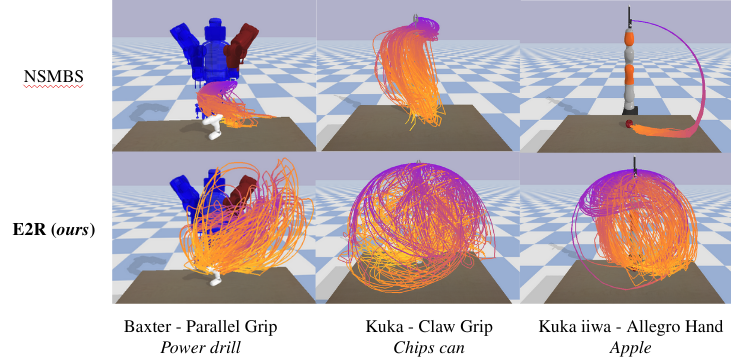}
  \caption{Diversity of grasping trajectories produced by E2R versus state-of-the-art \cite{morel2022automatic}. Each trajectory is plotted as a succession of end-effector positions in the Cartesian space. Color expresses temporality, from purple to yellow. Trajectories are plotted until the gripper touches the object. On each plot are drawn 250 randomly sampled trajectories from an output repertoire, obtained from runs on 3 different robots-gripper-object setups. While NSMBS's diversity is limited to local search around the first found trajectory, solutions generated by E2R are spread in the whole operational space.}
  \label{vis:div_approach_samples}
\end{figure}

Novelty-Search (NS) \cite{lehman2011abandoning} based methods face the exploration problem by driving the optimisation process with diversity instead of performance. Recent works show promising results on grasping \cite{liu2021pns}. In particular, \cite{morel2022automatic} introduces NSMBS, an algorithm that rely on multiple behaviorial descriptors (multi-BD) to generate diverse open-loop grasping policies on multiple robotic systems. By producing a large amount of diverse grasping trajectories, one can leverage such approach to build datasets that can be used to bootstrap learning of closed-loop grasping policies without strong limitations on the operational space. However, it appears that the diversity found by NSMBS is limited to a local search around the first found trajectory (Fig. \ref{vis:div_approach_samples}).

To tackle this issue, we introduce Explore-Refine-Regenerate (E2R), a new NS-based method that decomposes the task of grasping into two subtasks, approaching the object and grabbing it, within a unique evolutionary algorithm pipeline. That way, we leverage the inherent nature of grasping to make the generation of diverse and successful solutions easier, while keeping the simplicity of a standard NS with multi-BD framework. Contributions of this work are the following:

\BgThispage

\begin{figure*}[btp]
  \centering
  \includegraphics[width=\linewidth]{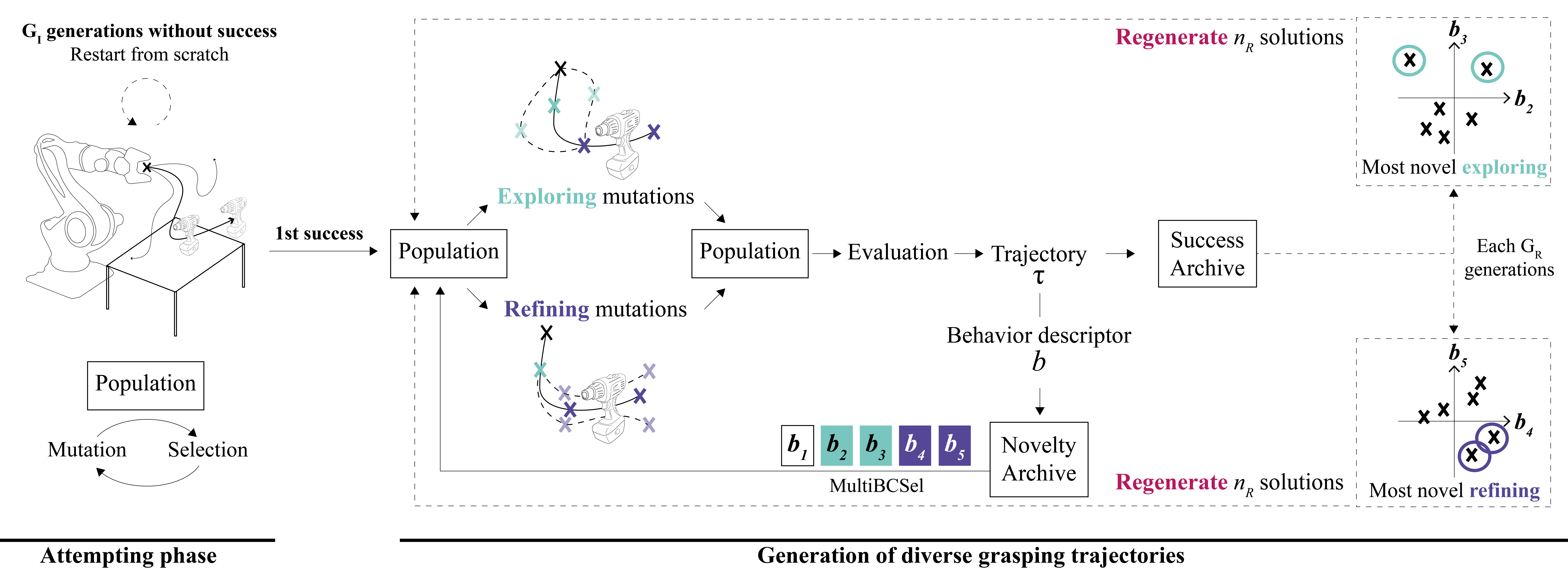}
  \caption{E2R overall principle. At first, we attempt to find a first successful grasp through a mutation-selection principle applied on a population of open-loop trajectories, favoring the most novel solutions \cite{lehman2011abandoning}. To find diverse policies, we then decompose the task into an approach ({\itshape explore}) and a prehension ({\itshape refine}) subtasks: mutation randomly focus on one of them, while selection is driven by behavior descriptors that distinguish diversity of approach ($b_2$, $b_3$) from diversity of prehension ($b_4$, $b_5$). The novelty archive is used as a long term memory of already discovered behaviors, pushing the search toward diversity on each descriptor (MultiBCSel \cite{morel2022automatic}). The population is frequenlty refilled with the most novel successful solutions on both tasks ({\itshape regenerate}). Output is a large and diverse repertoire of successful open-loop grasping policies.}
  \label{fig:diagram_overall_principle}
\end{figure*}

\begin{itemize}

\item We introduce a method that outperforms state-of-the-art for generating large repertoires of diverse grasping trajectories on several experimental setups, regarding: success ratio, diversity of both approach and prehension;
\item All the results are obtained using a common set of hyperparameters (except from the number of rollouts), regardless of robotic platform, gripper, and targeted object;
\item Preliminary experiments on a real robot show that some of the generated trajectories can successfully be transferred into the real world, paving the way for future work on the exploitation of the generated repertoires.

\end{itemize}
URL to the source code will be given if the paper is accepted. Supplementary materials and a video of experiments will be made available on the paper's webpage.

\section{RELATED WORKS}

\subsection{Learning to grasp}

After the early days of pure analytic-based approaches \cite{bicchi2000robotic}, machine learning got increasing importance since the beginning of the 21th century for robotic grasping \cite{kleeberger2020survey}. Imitation Learning and Gaussian Processes provide a convenient way to tackle the sparse reward problem, while being able to learn from a handful of examples \cite{chatzilygeroudis2019survey}. Inverse Reinforcement Learning proposes to make the agent infer a dense reward function from demonstrations \cite{horn2017quantifying}. Levine at al. \cite{levine2018learning} demonstrate that a Deep-RL approach can be bootstrapped with heuristic motor primitives to learn closed loop grasping policies. All those approaches are heavily constrained by the injected priors, which usually result into simple top-down movement with a parallel or suction gripper. Recently, Hindsight Experience Replay relies an a goal-conditioned RL paradigm to reduce the reward sparsity of a given task. To the best of our knowledge, this method has only shown results on real robot with parallel gripper and fixed orientation \cite{andrychowicz2017hindsight}.

\subsection{Generating repertoire of grasping policies}

Inspired by the success of supervised learning methods in the field of image processing \cite{deng2009imagenet}, many works tried to build large datasets of grasping trajectories \cite{mahler2016dex}\cite{fang2020graspnet}\cite{morrison2020egad}. But most of them simplify the problem as a grasping pose estimation task for parallel or suction grippers. While showing great results in closed loop, those datasets can hardly be applied on other kind of grippers, usually resulting into a top-down movement policies. Other works rely on motion capture on human demonstrations to build datasets of humanoïd-hand grasps \cite{saudabayev2018human}. Building those repertoires is very expensive as data is usually collected in the real world. In addition, each dataset is always specific to the arm and gripper used during data collection: learning to grasp on another kinematics or gripper would imply to rebuild the whole dataset.

\subsection{Leveraging diversity for robotics learning}

Quality-Diversity (QD) literature aims to couple diversity with quality during the optimisation process to get multiple high performing solutions \cite{pugh2016quality}\cite{cully2017quality}. Previous works rely on novelty criteria to get adaptation capabilities in navigation \cite{cully2015robots} or manipulation \cite{kim2021exploration} tasks. To address more complex problems, several works explored multiple behavioral descriptors (multi-BD) based methods \cite{koos2012transferability}\cite{pugh2016searching}. 

Grasping is not only a sparse reward but also a sparse behavioral problem, as most of the possible movements do not even touch the targeted object. To tackle this issue, NSMBS \cite{morel2022automatic} proposes a multi-BD approach that rely on hand-designed descriptors as proxies to generate successful grasps. This approach does not put heavy constraints on the operational space to generate successful grasps: the same hyperparameters work for grippers of different nature, including parallel and claw grips, but also humanoïd hands. The present work proposes to take a step further by significantly increasing the diversity of grasps, resulting into more exploitable repertoires.

\section{METHOD}

\subsection{Novelty Search}

Novelty Search based methods are evolutionary algorithms in which the optimization is driven by diversity instead of fitness. Originally introduced to tackle hard exploration problems \cite{lehman2011abandoning}, it also provides an efficient way to find diverse policies for a given objective \cite{kim2021exploration}.

Most of the following notations rely on the standard NS framework \cite{doncieux2019novelty}. Let $X$ be the space of parameters associated to a given {\itshape policy} $\pi$ (also called the {\itshape controller}). A policy maps an {\itshape observation} (e.g. position of a robotic manipulator) returned by the {\itshape environment} we are operating in (e.g. a robot, a table and an object submitted to fixed physical dynamics) to a specific {\itshape action} (e.g. motors actuation). Let $x \in X$ be an {\itshape individual}. With NS, we make a {\itshape population} of individuals $x_i \in X$, $i \in \{1, ..., \mu\}$ evolve to maximize their diversity with respect to a given task. To do so, we evaluate the parametrized policies $\pi_{x_i}$ on a Markov Decision Process (MDP) \cite{howard1960dynamic} $\cal{M}$, corresponding to the task's environment. We then obtain a set of trajectories $\tau_i \in S_\tau$, defined as a sequence of states and actions for each step of an {\itshape episode}: a {\itshape rollout} (i.e. an evaluation) corresponds to a fixed number of steps.

$S_{\tau}$ is usually in high dimension (e.g. each of the $T$ joint parameters of a robotic system, where $T$ is the length of an episode). To evaluate trajectories' novelty, we project $\tau$ from $S_{\tau}$ to a space $\cal{B}$ called {\itshape behavior space}, that verifies $dim({\cal{B}}) << dim(S_{\tau})$. This projection results into a {\itshape behavior descriptor} $b \in \cal{B}$ (noted $b(x)$, and also called {\itshape behavior characterization} \cite{pugh2016searching}) associated to the trajectory. In this work, we consider the NS framework in which each episode starts from the same state. $\cal{M}$ is assumed deterministic.

During the exploration process, two trajectories $\tau_{1}$ and $\tau_{2}$ (and, by extension, the two corresponding individuals $x_{1}$ and $x_{2}$) are considered similar if $b_{1} \approx b_{2}$. Contrary to standard fitness-based optimisation methods \cite{eiben2003introduction}, NS leverages a criterion of novelty to drive the exploration process. The novelty $\rho$ is measured as the average distance of an individual $x$ to its closest neighbors within the behavior space, such that: 
\begin{equation}
\label{eq:novelty}
\rho(x) = \frac{1}{k} \sum_{i=1}^{k} dist(b_{x}, b_{\mu_i})
\end{equation}
where $\mu_i$ are the k closest neighbors in $\cal{B}$ within a {\itshape reference set} defined as the union of an {\itshape archive} of previously generated individuals and the current population. Note that $dist$ is usually the euclidean distance \cite{morel2022automatic}\cite{koos2012transferability}. 

Overall, NS can be described as assuming that a projection from $X$ to $\cal{B}$ exists (called the {\itshape behavior function}), and making the design choices that allow the exploration of $\cal{B}$ by searching in $X$ through a novelty-driven divergent search \cite{doncieux2019novelty}. In this work, the exploration of $\cal{B}$ is used as a proxy to generate solutions which verify a success condition (i.e. grasping an object), while being as diverse as possible in $S_{\tau}$.

\subsection{Intuition}

E2R is designed to make the generation of diverse grasping trajectories easier. To do so, our main concern is to make the behavioral landscape ${\cal{B}}: x \mapsto b$ as smooth as possible, that is, making sure that a small step in $X$ will result in a small step in $\cal{B}$. On the contrary, we would like the trajectory landscape $S_{\tau}: x \mapsto \tau$ to be less smooth as possible, so we could maximize diversity of the generated trajectories. For a given $x$ and a small variation $\delta x$, the problem can be formulated as making the design choices that verify:
\begin{equation}
\label{eq:optim}
\left\{\begin{matrix}
\underset{\epsilon_{\cal{B}}}{\min} \left| {\cal{B}}(x) + {\cal{B}}(x+\delta x) \right| < \epsilon_{\cal{B}} \\
\underset{\epsilon_{S_\tau}}{\max} \left| {S_\tau}(x) + {S_\tau}(x+\delta x) \right| < \epsilon_{S_\tau}
\end{matrix}\right. 
\end{equation}
Under those conditions, a small step in $X$ from a successful solution $x_s$ is likely to result into another successful solution, which can be different from a trajectory perspective.

The nature of grasping task makes this optimisation problem really difficult. It is widely admitted in the field that grasping can be considered as a succession of 3 subtasks \cite{newbury2022deep}: approaching the object, applying forces on it (referred in the present work as the {\itshape prehension} phase), and verifying the outcome of the grasp. The most problematic subtask is the interaction between the object and the grip, submitted to strong discontinuities: a small variation of gripper's orientation or applied forces might lead to dramatically different outcomes.

Hierarchical Reinforcement Learning (HRL) decomposes decision making problems into smaller ones to make the learning process easier \cite{pateria2021hierarchical}. Inspired by this paradigm, we made the assumption that generating diverse trajectories of grasping is easier if we consider the task of approaching and grabbing separately. While approaching is a navigation task, grabbing belongs to manipulation tasks, as it implies interaction between the gripper and the object. Considering both tasks as a single one makes the navigation task more complicated than it is in practice, as it brings unpredictability inherent to the discontinuity of manipulation to an easily predicted task.

By considering grasping as the succession of an approach and a prehension tasks, we can do local search from a successful individual in such way that limits the impact of the variation on the resulting behavior. Indeed, modifying the approach trajectory while applying an already found prehension strategy is likely to result into a successful grasp. Similarly, keeping an approach trajectory unchanged in a deterministic environment will eventually lead to the state from which a successful prehension has already be found, so that a small trajectory variation from that point has significant chances to produce another successful solution.

\subsection{How to divide the task without actually doing it}

Decoupling approach from prehension requires: 1) to explicitly push the exploration process toward diversity from both subtasks; 2) to do steps in the search space that would affect one of the subtask, with limited impact on the other one. 

Let $T$ be the fixed length of episodes, and $t_{touch}$ be the first time step in which the robot touches the object. To address the first requirement, E2R relies on the following descriptors: 

\begin{itemize}

\item $b_1$: Object's position at $t=T$. It guides the search toward solutions that make the object move, and eventually lift it.
\item $b_2$: End-effector's position at $t=t_{touch}$. Stimulates the emergence of diverse grasp points on the object.
\item $b_3$: End-effector's orientation at $t=t_{touch}$. Pushes the exploration process to try to grasp the object from different angles.
\item $b_4$: End-effector's position at $t=T/2$. Favors diversity in approach trajectories.
\item $b_5$: End-effector's orientation at $t=T/2$. Incites agents to approach the object from different angles.

\end{itemize}
While $b_1$ pushes the exploration toward successful solutions, ($b_2$, $b_3$) and ($b_4$, $b_5$) are proxies respectively dedicated to approach or prehension.

The second requirement is addressed by leveraging the structure of the task, the nature of the controller, and the mutation operator (i.e. the function which applies optimisation steps to individuals in $X$). The task is considered as an episode-based MDP \cite{howard1960dynamic} of length $T$. Controllers are encoded as a 3 way-points open-loop policy, which are used as reference for a polynomial interpolated trajectory. Each individual $x$ consists of the concatenation of 3 joint positions to reach during the trajectory, plus a scalar that defines the step of the episode in which the constant-speed closure of the gripper is to be initiated. The way points are uniformly distributed throughout the episode. To avoid issues caused by non-null speed grasp \cite{meszaros2022learning}, the position of the robotic arm is fixed during the gripper closure. 

The 3-way points controller is designed to implicitly encode the two subtasks involved in grasping. The fixed length of episodes and the limits imposed by systems kinematics make successful solutions fit a convenient structure: the first way point is the unique driver for diversity of approach, the second is a proxy to get the grip ready to grab the object, and make the last point the target to reach in order to verify the outcome of the grasp. Such a controller favors the behavioral landscape smoothness: one can modify the first way-point without compromising the grasp as long as the other points do not change, while keeping the approach points similar to successful solutions is likely to lead to slightly different solutions that still grasp the object.  

Lastly, we defined a mutation operator that decouple approach from prehension in the search space. Each time we want an individual to find new approach trajectories, we set a large variance on the mutation of its first way point, and set a near-zero variance on the evolution of the second and third ones. Similarly, each time we want an individual to find other way to grasp the object, we set a small variance on the first point mutation, and a large variance on the mutation of the second and third way points.

E2R's principle is summarized in Fig. \ref{fig:diagram_overall_principle}. To generate a diverse repertoire of grasping trajectories, we rely on Multi-BD based method similarly to \cite{morel2022automatic}, while making the mutation-selection process focus on either approach or prehension. At first, the agent must discover a pioneer successful trajectory ({\itshape attempt phase}). As soon as a first successful solution is found, we can rely on local search to find close-but-different grasping trajectories by applying either the {\itshape exploring} mutation or refining {\itshape refining} mutation on individuals. A {\itshape Regeneration} module drives the trade-off between exploration and exploitation, by frequently resetting the population with the most novel successful solutions for the two subtasks independently.

\begin{algorithm}
\caption{E2R}\label{alg:e2r}
\hspace*{\algorithmicindent} \textbf{Input:} $\mu, \lambda, G, p_e, p_r, G_I, G_R, n_a, k$ \\
\hspace*{\algorithmicindent} \textbf{Output:} $a_s$
\begin{algorithmic}
\State $x_p \gets initPop(\mu)$
\State $a, a_s \gets \emptyset, \emptyset$
\State \text{{\bf for} $g$ {\bf in} $1,...,G$ {\bf do}}
    \State \hskip1.0em \text{{\bf if} $card(a_s) = 0$ \text{\bf and} $isImpatienceGen(g,G_I)$ {\bf do}} \\
        \hskip2.0em $x_p \gets initPop(\mu)$ \\
    \hskip1.0em \text{{\bf end if}}
    \State \hskip1.0em \text{{\bf if} $isRegenerateGen(g,G_R)$ {\bf do}} \\
        \hskip2.0em $x_p \gets regenerate(\lambda,a_s)$ \\
    \hskip1.0em \text{{\bf end if}} \\
    \hskip1.0em $x_o \gets randomSample(x_p, \lambda)$ \\
    \hskip1.0em $x_o \gets mutateER(x_o, p_e, p_r)$ \\
    \hskip1.0em $evaluate(x_o)$\\
    \hskip1.0em $a_s.add(getSuccesses(x_o))$\\
    \hskip1.0em $x_{po}$, $x_{poa} \gets x_p \cup x_o, x_p \cup x_o \cup a$ \\
    \hskip1.0em $updateNovelty(x_{po}, x_{poa}, k)$\\
    \hskip1.0em $a.add(randomSample(x_o, n_a))$\\
    \hskip1.0em $x_p \gets multiBCSel(x_{po},\mu)$
\State \text{{\bf end for}}

\end{algorithmic}
\end{algorithm}

\subsection{Algorithm}

Algorithm \ref{alg:e2r} describes E2R's overall procedure. The population $x_p$ is initialized with $\mu$ individuals randomly sampled from a uniform distribution, and evaluated on the domain ($initPop$). At each generation, we first select $\lambda$ individuals from $x_p$ to form the offspring $x_o$. They are then mutated, and evaluated on the domain. Successful individuals are added to the output success archive $a_s$. The novelty of individuals from both the population and the offspring ($x_{po}$) are updated based on the reference set ($x_{poa}$) by applying a k-nearest-neighbour method on the behavior descriptors. We then randomly sample $n_a$ individuals from offspring and add them to the novelty archive $a$. The next generation population is selected based on their novelty with respect to each behavior descriptor, following the $multiBCSel$ routine proposed in \cite{morel2022automatic}.

During the attempting phase, the evolutionary process might get stuck: if many of the first attempting trajectories fail, the archive is saturated with descriptors which are now considered as unpromising, even if they are good stepping stones toward successful grasps. To prevent the search from being stuck, an {\itshape impatience} mechanism has been implemented such that the evolutionary process is restarted every $G_I$ generations as long as no successful solution has been found.

Each $G_R$ generation, at most $\mu$ successful individuals are selected from $a_s$, and added to the population ($regenerate$). Note that we select $n_R$ most novel individuals for respectively exploring or refining, such that:
\begin{equation}
\label{eq:regen}
n_R =  \frac{\min(\mu, card(a_s))}{2}
\end{equation}
That way, the population is equally composed by solutions which are respectively the most likely to mutate into undiscovered way to reach or to grab the object.

\section{EXPERIMENTAL SETUP}

\subsection{Environments}

Experiments are conducted on the simulator Pybullet \cite{coumans2016pybullet}. Methods are compared on 3 robotics platforms: a Baxter with a parallel grip, a Kuka with a claw grip, and a Kuka iiwa with an Allegro hand. Baxter and Kukas’ arms are 7 degrees-of-freedom manipulators. The experiments are conducted on a subset of 5 diverse rigid objects from the YCB dataset \cite{calli2015benchmarking}: {\itshape apple}, {\itshape chips can}, {\itshape gelatin box}, {\itshape mug}, and {\itshape power drill}.

\subsection{Algorithms}
  
In this study, we compare 4 algorithms on each of the 15 (robot, object) pair: NS, Random, NSMBS and E2R. To the best our knowledge, NSMBS \cite{morel2022automatic} is the current best platform-agnostic method that can generate large dataset of grasping behaviors. A standard NS \cite{lehman2011abandoning} and an evolutionary process with pure-random selection are compared as baselines. All of those methods are using the same 3-way points architecture of policy. Note that NS was evaluated on all E2R's behavioral descriptors, concatenated as a single vector. Each non-eligible component is set to $0$ by convention.

Experimental hyperparameters are the following: $\mu=100$, $\lambda=50$, $p_e=50$, $p_e=p_r=0.5$, $G_I=500$, $G_R=10$, $n_A=10$, $k=15$. For fair comparisons, we fixed a maximal number of rollouts $N_{rt}$. Methods are compared in simulation with $N_{rt}=200k$ for Baxter, and $N_{rt}=400k$ for both Kuka and Kuka iiwa with Allegro hand. Sim2real transferability are evaluated with repertoires generated under $N_{rt}=25k$ rollouts.

\begin{figure}[h]
  \centering
  \includegraphics[width=\linewidth]{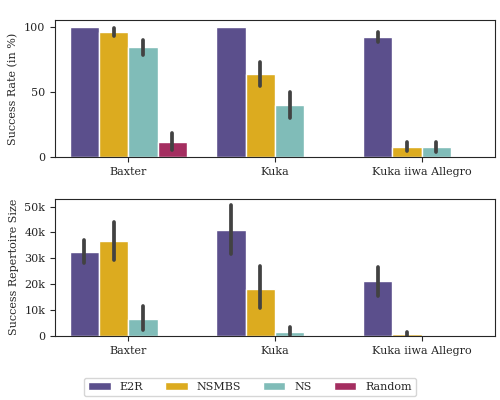}
  \caption{Success ratio and size of the output repertoire of grasping trajectories on evaluated methods. Averaged over 5 seeds on all the evaluated objects. Error bars are $0.95$ confidence interval. E2R outperforms all other methods in success rate on the evaluated setups. The number of generated solutions make E2R and NSMBS the two most promising methods to generate diverse grasping trajectories.}
  \label{fig:scsratio_hist}
\end{figure}

\begin{figure}[btp]
  \centering
  \includegraphics[width=\linewidth]{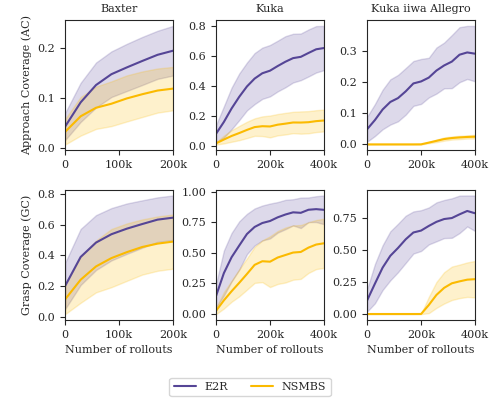}
  \caption{Evolution of success repertoire's diversity throughout the evolutionary process. Averaged over 5 seeds on all the evaluated objects. The bands express the standard deviation. E2R generates more diverse trajectories on the evaluated setups, regarding both approach and prehension ($p < 10^{-3}$).}
  \label{fig:diversity_lp}
\end{figure}

\subsection{Metrics}
  
The present work relies on 3 sets of metrics, that only take into account successful trajectories within the deterministic simulated environment. Firstly, we compare each method on their ability to generate at least one success during a run ({\itshape Success rate}), as well as on the number of successful grasps found during the run ({\itshape Success Repertoire size}). Secondly, we compute the {\itshape Approach Coverage} (AC), and {\itshape Grasp Coverage} (GC) by respectively discretizing the operational space and the object's surface. AC expresses the ratio of discrete cells of the operational space that have been occupied by the end-effector at least once along a trajectory, while GC expresses the number of discretized voxels of the object surface which contains the first object-gripper contact point for at least one of the trajectories. The operational space is discretized so that each cell describes a volume of 4 cm³, and each voxel on the surface of objects corresponds to a volume of 1 cm³. Lastly, we estimate the {\itshape Sim2real success ratio} of a generated repertoire as its ratio of policies which lead to successful and robust grasps in reality (i.e. the object is lifted over the table, and does not fall off the gripper if we shake it). Each sampled trajectory is deployed 3 times, and is considered as a success if at least one trial leads to a successful grasp.

\section{RESULTS AND DISCUSSION}

Fig. \ref{fig:scsratio_hist} presents the obtained successful run ratios for all the compared methods. Overall, E2R outperforms all the evaluated methods regarding success ratios ($p<3*10^{-2}$, averaged on all objects): On each seed, E2R generates at least one successful solution on Baxter and Kuka with a claw gripper, while the most competitive solution (NSMBS) obtains respectively 96\% and 64\% of success rate. On the most difficult task (Kuka iiwa with Allegro hand), E2R is clearly heads above the competition with 92\% of success rate, while none of the compared algorithms exceeds 8\%. Random search got only 12\% of success rate on Baxter, while being unable to generate a single successful solution on the other robots. NS performs way better than Random regarding success rate (84\% on Baxter, 40\% on kuka, 8\% on kuka iiwa allegro), but is not competitive with E2R and NSMBS regarding the number of generated trajectories. On Baxter, which appears to be the easier of the evaluated setups, NS generates less that 7k solutions, while E2R and NSMBS respectively generates about 32k and 36k solutions. Those results suggest that: 1) E2R is the best method among the compared ones to produce successful grasping trajectories with minimal number of runs, making it the most affordable method regarding the computation cost; 2) NSMBS and E2R are the most promising candidates on the targeted task, deserving further investigation on the nature of produced solutions regarding diversity.

\begin{figure}[btp]
  \centering
  \includegraphics[width=\linewidth]{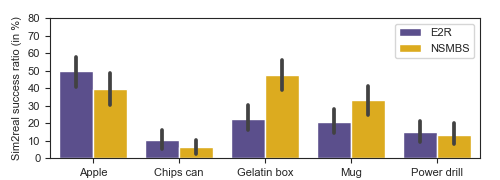}
  \caption{Sim2real transferability rate of randomly sampled trajectories from success repertoires on a real Baxter robot. Averaged on 3 seeds. Error bars are $0.95$ confidence interval. The large diversity of E2R's generated solutions does not compromise their exploitability on a real robot.}
  \label{fig:sim2real_hist}
\end{figure}

Fig. \ref{fig:diversity_lp} shows the evolution of success repertoire diversity throughout the search. E2R get higher diversity coverages than NSMBS on each of the tested robot-object pair ($p < 10^{-3}$). Even in tasks in which NSMBS got a high success rate and generates a large output repertoire (Baxter), E2R obtains a higher AC and GC on average on all the evaluated objects. Consistency of experimental results beyond robots' kinematics, grippers and objects suggests that E2R is the current best method to generate a large diversity of trajectories, regarding both the approach and the prehension parts of grasping. On Fig. \ref{vis:div_approach_samples} are drawn 250 randomly sampled trajectories from some of the E2R or NSMBS output repertoires. This visualisation confirms the obtained results, as the grasping trajectories generated by E2R cover a wider region of the operational space than those generated by the current state-of-the-art.


Fig. \ref{fig:sim2real_hist} shows the obtained sim2real transfer ratios obtained by deploying randomly sampled grasping trajectories on a real Baxter robot. Even with such a rough selection process, successful transfers have been observed for both methods and objects. Objects which are subject to model inconsistencies (e.g. erroneous mass distribution) lead to the lower transferability rates. For example, the chips can led to about 12\% and 7\% of success respectively for E2R and NSMBS. On the contrary, the simulated model of the apple does not seem to be subject to such weaknesses and result into high transferability rate on average (about 37\% for NSMBS and 54\% for E2R).

\begin{figure}[btp]
  \centering
  \includegraphics[width=\linewidth]{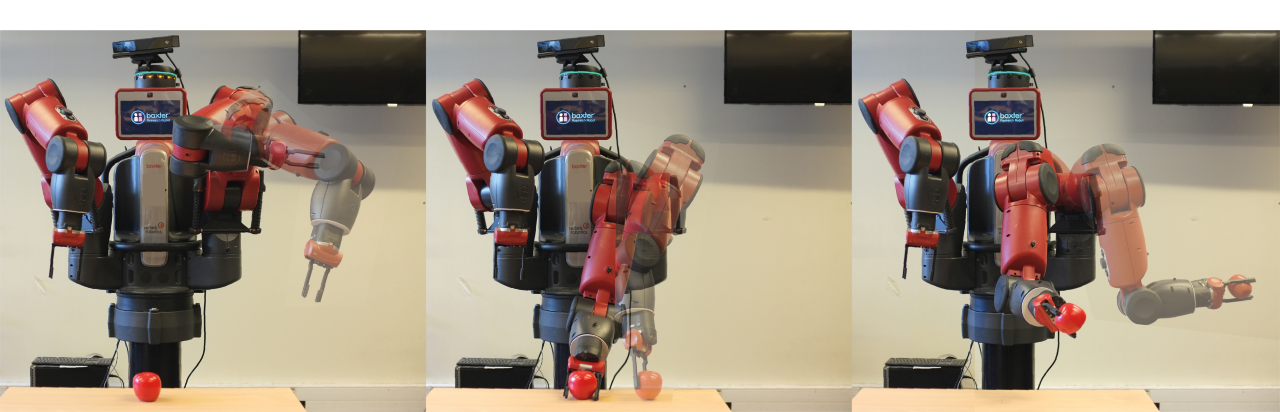}
  \caption{Example of successfully transferred grasps generated by our method. Sampled trajectories are significantly diverse regarding both the approach phase and application of forces on the object.}
  \label{fig:real_bx_photo}
\end{figure}

Generating large repertoire of diverse trajectories is not meant to be used by randomly sampling trajectories to deploy: favoring diversity in simulation is likely to push generation of solutions which exploit simulation's imperfections, making them inefficient in reality \cite{collins2019quantifying}. Still, this preliminary experiment shows that even when doing random sampling, some of the trajectories lead to successful grasps in reality (Fig. \ref{fig:real_bx_photo}). On a large number of trials, we expect this experiment to result into a higher variance of sim2real success ratio for NSMBS than E2R, considering the fact that NSMBS is constrained by the first found successful grasp, around which the method do local exploration. If the first found grasp is a physically meaningful grasp (i.e. matching analytics criteria  \cite{nguyen1988constructing}), many of the other solutions are likely to transfer. On the contrary, if the first found trajectory is a simulation exploit, other solutions are likely to fail when being  deployed on a real robot. As an example, 2 different NSMBS seeds on the mug respectively led to 2\% and 60\% of sim2real success ratios. By generating much more diversity, E2R is more likely to generate a minimal number of transferable solutions. The drawback of such a large exploration is that E2R is also more likely to find more simulation exploits, requiring to carefully select the most promising solutions for sim2real transfer. Though, we believe that a repertoire which efficiently cover the object's affordance can be more easily exploited than an subset of grasping trajectories that focus on a small part of the object.

\section{CONCLUSIONS}

The design choices of the E2R mutation-selection mechanism make the search to focus on either approach (i.e. reaching the object) or prehension (i.e. applying forces and torques on its surface), while keeping the simplicity of a standard NS pipeline. Our method generates repertoires of grasping trajectories which are significantly more diverse than the current state-of-the-art on all the evaluated setups. Experiments on a real Baxter show that the obtained solutions can successfully be transferred into the real world. 


\section*{ACKNOWLEDGMENT}

This work was supported by Sorbonne Center of Artificial Intelligence (SCAI), and by the project Learn2Grasp. This work used HPC resources from GENCI-IDRIS (Grant 20XX-AD011013516).

\bibliographystyle{IEEEtran}
\bibliography{refs}

\addtolength{\textheight}{-12cm}   




\end{document}